\documentclass[conference]{IEEEtran}

\usepackage[utf8]{inputenc}
\usepackage[T1]{fontenc}
\usepackage{microtype}

\usepackage[keeplastbox,ancient]{flushend}

\usepackage{booktabs}
\usepackage{graphicx}
\usepackage{url}

\usepackage{enumitem} 
\setlist{nosep}
\setlist[itemize,1]{leftmargin=\dimexpr 1em}

\usepackage{csquotes}

\makeatletter
\def\ps@IEEEtitlepagestyle{%
  \def\@oddfoot{\mycopyrightnotice}%
  \def\@evenfoot{}%
}
\def\mycopyrightnotice{%
  {\begin{minipage}{\textwidth}
  \footnotesize \copyright 2019 IEEE. Personal use of this material is permitted. Permission from IEEE must be obtained for all other uses, in any current or future media, including reprinting\slash republishing this material for advertising or promotional purposes, creating new collective works, for resale or redistribution to servers or lists, or reuse of any copyrighted component of this work in other works.
  \end{minipage}
  }
  \gdef\mycopyrightnotice{}
}

\begin{document}

\title{Requirements Engineering for Machine Learning: Perspectives from Data Scientists}

\author{\IEEEauthorblockN{Andreas Vogelsang}
\IEEEauthorblockA{Technische Universit\"at Berlin\\
Berlin, Germany\\
andreas.vogelsang@tu-berlin.de}
\and
\IEEEauthorblockN{Markus Borg}
\IEEEauthorblockA{RISE Research Institutes of Sweden AB\\
Lund, Sweden\\
markus.borg@ri.se}
}


\maketitle

\begin{abstract}
Machine learning (ML) is used increasingly in real-world applications.
In this paper, we describe our ongoing endeavor to define characteristics and challenges unique to Requirements Engineering (RE) for ML-based systems. 
As a first step, we interviewed four data scientists to understand how ML experts approach elicitation, specification, and assurance of requirements and expectations. 
The results show that changes in the development paradigm, i.e., from coding to training, also demands changes in RE. We conclude that development of ML systems demands requirements engineers to:  
(1) understand ML performance measures to state good functional requirements, (2) be aware of new quality requirements such as explainability, freedom from discrimination, or specific legal requirements, and (3) integrate ML specifics in the RE process.
Our study provides a first contribution towards an RE methodology for ML systems.
\end{abstract}


\section{Introduction}
Machine Learning (ML) has gained much attention in recent years and accomplished major technical breakthroughs. ML success stories include image recognition, natural language processing, and beating humans in complex games~\cite{Khomh18}. The key enablers for the ongoing ML disruption are based on the combination of the computational power of modern processors (especially GPUs), the availability of large amounts of data, and the accessibility of these complex technologies by rather simple-to-use frameworks and libraries. As a result, many companies use ML to improve their products and processes.

Whether the increasing amount of ML in contemporary software solutions demands requirements engineers to adapt their work is an open question. One might argue that ML is just another technology, i.e., it should not have any influence on the requirements. On the other hand, ML engineering constitutes a paradigm shift compared to conventional software engineering. Andrej Karpathy, Director of AI at Tesla, even refers to the new era as \enquote{Software 2.0} -- no longer does all behavior emerge from a set of manually coded rules. Instead, most ML approaches generate rules based on a set of examples (training data) and a specific fitness function. In addition, a recent survey suggests that Requirements Engineering (RE) is the most difficult activity for the development of ML-based systems~\cite{Ishikawa19}. 

In this paper, we share results that indicate that RE must evolve to match the specifics of ML systems. Currently, most decisions in the development of ML systems are made by data scientists. These decisions include the definition of the fitness functions, the selection and preparation of data, and the quality assurance. However, these decisions should be based on an understanding of the business domain and the stakeholder needs. From our perspective, this falls into the profession of a requirements engineer. 


We conducted interviews with four data scientists to explore their perceptions on RE. The interviews covered specific requirements for ML systems, challenges involved in RE for ML, and how the RE process needs to evolve. Our main findings are that requirements engineers need to be aware of new requirements types introduced by the ML paradigm, e.g., explainability and freedom from discrimination, and they need to understand quantitative ML measures to specify good functional requirements. We elaborate on our results in Sections~\ref{sec:reqts} and~\ref{sec:process}, after having presenting background in Section~\ref{sec:bg}, and the study design in Section~\ref{sec:method}.

\section{Background and Related Work} \label{sec:bg}
\subsection{Machine Learning and Software Engineering}
ML is the practice of getting computers to act without being explicitly programmed, organized in three main types. \textit{Supervised learning} finds mappings between input-output pairs based on labelled examples with the correct answer. This type of ML dominates industrial applications today, but requires high-quality data. \textit{Unsupervised learning} identifies patterns in data without access to any labels. A typical application of this type of ML is clustering. \textit{Reinforcement learning} relies on a reward signal to quantify the ML performance, inspired by the carrot-and-stick principle. Video games have been the primary target for this approach, but software engineering examples exist, e.g., software testing~\cite{moghadam2019machine} and self-adaptive systems \cite{wu2018using}.


Using ML to learn system features is fundamentally different from manually implementing them in source code. The predictive power is embedded in the training data, harnessed using a limited number of function calls. State-of-the-art supervised learning use backpropagation to fit millions of parameter weights representing information propagation in deep neural networks, resulting in an ML model. Compared to human-readable source code, ML models are opaque constructs that are intrinsically hard to interpret. Standard practices such as source code reviews and exhaustive coverage testing are not applicable to ML models~\cite{borg2018safely} -- in this paper, we show that also RE must evolve to meet the needs of ML systems.

\subsection{Requirements Engineering for ML}
While there are approaches on how to use ML to improve RE tasks (e.g., prioritization~\cite{Perini13}, model extraction~\cite{Arora19,Pudlitz19}, categorization~\cite{WinklerV16}, app store analytics~\cite{Maalej16}), there is not much work on RE for ML systems. Although current reference processes for data mining such as the Knowledge Discovery in Databases (KDD) process~\cite{Maimon10} or the Cross-industry standard process for data mining (CRISP-DM)~\cite{Shearer00} have initial steps that are called ``business understanding'' or ``Developing an understanding of the application domain'', neither the RE nor the ML community caught up on that and detailed what these steps demand.
In a recent survey, Ishikawa and Yoshioka~\cite{Ishikawa19} reported that RE was listed as the most difficult activity for the development of ML-based systems: 
\enquote{The dominant concerns [\ldots] pertain to decision making with the customers. In the conventional setting, this activity involved requirements analysis and specification in the initial phase and an acceptance inspection in the final phase. This activity flow is not possible when working with ML-based systems due to the impossibility of prior estimation or assurance of achievable accuracy.}
Horkoff states challenges and research directions for non-functional requirements for ML systems~\cite{Horkoff19}. She argues that there is no unified collection or consideration of many NFRs for ML, including a consideration of ML-specific quality trade-off data. Current work consists of only individual considerations of specific quality trade-offs, e.g., privacy vs. processing time. As one possible research direction, she suggests asking ML experts about non-functional requirements, which is what we do in our paper.

\subsection{RE for Other Data-Driven Paradigms}
Self-Adaptive Systems (SAS) continuously collect data to monitor their run-time behavior~\cite{mahdavi2017classification}. Based on the data, the systems decide whether adaption is required due to e.g. changes in the operational environment, system faults, or changing requirements. Self-adaptation is an approach to tackle limited knowledge at design-time, to instead deal with uncertainty at runtime when more data is available. Example applications include cloud computing, climate control in smart buildings, and firewalls that adapt protection mechanisms to block cyberattacks.

RE for SAS is challenging as the requirements analysis is inherently incomplete~\cite{cheng2009software}. Thus, many decisions related to RE must be postponed to runtime and rely on data to adapt to various circumstances. Kephart and Chess propose a Monitor-Analyze-Plan-Execute loop to realize the adaptation by monitoring requirements satisfaction and making changes based on the knowledge modelled at requirements-time~\cite{kephart2003vision}. Morandini et al. suggest the Tropos4AS framework to facilitate engineering of SAS~\cite{Morandini17}. Tropos4AS supports the requirements analysis by providing modelling features for knowledge and decision criteria needed by systems to autonomously adapt to changes, i.e., a goal model, an environment model, and a failure model.

Both ML systems and SAS are data-intensive, but SAS are not necessarily trained on data -- instead their adaptation can be implemented in conventional source code. (Supervised) ML systems, on the other hand, are trained on data -- but might not be self-adaptive, since they do not necessarily perform continuous monitoring of their operational environments. A supervised ML system might be trained once on data before deployment, but never again retrained on new data. Such a system can be referred to as a \textit{trained} but not a \textit{learning} ML system. However, there is obviously potential in the combination of ML and self-adaptation (a recommended direction for future work by Morandini et al.~\cite{Morandini17}), but we argue that RE for ML involves characteristics that are not covered by RE for SAS.

(Big) Data Analytics is another field closely related to ML. Nalchigar and Yu~\cite{Nalchigar18} propose a modeling framework for requirements analysis and design of data analytics systems.
It consists of three complementary modeling views: business view, analytics design view, and data preparation view. These views are linked together to connect enterprise strategies to analytics algorithms and to data preparation activities. Especially the analytics design view and the data preparation view may be well suited to support requirements derivation also for ML components. Liu et al.~\cite{Liu16} explore steps that healthcare organizations take to elicit data analytics requirements, to specify functional requirements for using data to improve business and clinical outcomes.

\section{Study Design} \label{sec:method}
\noindent We were interested in the following research questions:
\begin{itemize}
    \item \textbf{RQ1:} How do data scientists elicit, document, and analyze requirements for ML systems?
\item\textbf{RQ2:} What processes do data scientists follow and which parts relate to requirements?
\item\textbf{RQ3:} What are requirements-related challenges that data scientists face?
\end{itemize}

\subsection{Research Method}
We employed a qualitative approach, using semi-structured interviews. Semi-structured interviews employ an interview guide with questions, but allow the order of questions to vary to fit the natural flow of the conversation.
Our interview questions\footnote{Interview guide: \url{https://doi.org/10.6084/m9.figshare.8067593.v1}} took an exploratory approach. We asked respondents about their background, typical projects that they are involved in, and the role of ML therein. Afterward, we specifically asked for ways how they elicit, communicate, document, and test expectations and requirements for their applications. 
In the final part of the interviews, we went through a list of RE best practices and asked the participants about best practices they use in their projects.

\subsection{Study Subjects}
Since ML is a rather recent and complex technology, most ML-based systems are driven by people who are experts in the technology. They are the people who currently prepare the data, make design decisions, and finally evaluate the performance of their systems. Therefore, we selected people from the field of data science as subjects.
We interviewed four data scientists with each interview lasting approximately one hour. For confidentiality, respondents are kept anonymous and referred to with running ids P1--P4. P1 and P2 are working in research and have 3--4 years of experience as data scientists. P1 does research on data integration, data profiling, and data discovery. P2 focuses on data engineering. P3 and P4 work in industry. P3 has over 20 years of working experience as data scientist\footnote{Although not called \emph{data science} 20 years ago, P3 was involved in projects where tools made decisions based on rules inferred from existing data.}. Currently, he is in the mobility domain where he works on data-driven systems to increase user experience in cars. P4 works in finance for 4 years and develops all kinds of data-driven applications that are related to creditworthiness and loan loss risks.
All interviews were conducted by the first author and were recorded and transcribed. 

\subsection{Data Analysis} \label{sec:analysis}
We relied on thematic coding~\cite{Gibbs08} in a collaborative setting. Each author started with coding one interview and afterward reviewed and validated the codes in the interview transcript of the other author. The results and lessons learned were discussed in a meeting. Then, each author coded a second interview transcript, which was afterward, again, validated by the other author. Statements in the transcripts were assigned one or more codes. 
In particular, we used a mixture of a priori coding based on our research questions, and emergent coding starting from any mention of requirements. 
We then combined data from all transcripts in a meeting to ensure that we do cover the full data in our synthesis of findings. 
Quotes have been translated from the respondents' native language to English and edited for readability. Colloquialisms have been kept to convey the tone of the conversation and to reflect the informal nature of the interview setting.

\subsection{Threats to Validity}
Maxwell~\cite{Maxwell12} identified five threats to validity in qualitative research that also apply to our study design. \emph{Descriptive validity} refers to the threat that an interviewer does not collect all relevant data during an interview. To mitigate this threat, we recorded the interviews on tape. We annotated the transcripts with pointers to the respective positions in the recording to be able to trace back the original conversation. \emph{Interpretation validity} refers to the possibility of misunderstandings between interviewees and the researchers. To minimize this risk, the study goal was explained to the participants prior to the interview. Steps taken to improve the reliability of the interview guide included a review. \emph{Researcher bias} and \emph{theory validity} are threats that refer to the bias of the researcher to interpret the interviews in a way that serves his or her goals or initial theory. Since this is our first study in this area, we do not have a specific RE methodology that we would like to promote. That means, we were very much open to the outcomes of the interviews. Furthermore, we tried to lower researcher bias by cross-validating the annotated codes by the two authors, which has also not collaborated before on this topic. \emph{Reactivity} refers to the threat that interviewees behave differently because of the presence of the interviewer.  Getting rid of reactivity is not possible, however we are aware of it and the way it may influence what is being observed.

\section{Challenges and Requirements for ML Systems} \label{sec:reqts}

\subsection{Quantitative Targets a.k.a. Functional Requirements}
Just like for conventional software products, it is hard to satisfy a client without making expectations explicit. P3 stressed the importance of quality targets for ML models: \enquote{For a successful project, the requirements must be clear. Especially the evaluation metric must be specified}.
In ML systems, the quality of the resulting predictions can be considered a functional requirement (P3: \enquote{I consider predictive power as functional requirement.}) Our interviewees mentioned measures to quantify the predictive power including accuracy, precision, and recall. P3: \enquote{The customers don't understand the performance measures.} Selecting and interpreting performance measures appropriately is crucial for the acceptance of ML systems. P1: \enquote{Consider an  example of detecting errors in a dataset. Usually, you have less errors than non-errors. The data is imbalanced. If you have 10\% errors, a trivial classifier  always predicting \emph{no error} has an accuracy of 90\%. Recall would be a better measure here.} 
P1 stressed that data scientists need skills to help a client set reasonable targets, including domain understanding, statistics, and computer science. We believe that these skills will also be needed for successful RE for ML. 
P3 had positive experiences with a measure that is not so common: \enquote{Lift\footnote{Lift is the factor by which the prediction of a model is more accurate than a random choice.} is often a good measure. [\ldots] It is better than precision because most people don't understand that precision depends on the a-priori distribution of the data. If you want to detect women between 20 and 40 and they make 15\% of the population, a model with a precision of 60\% has a lift of 4, which is good.} 

Quantification of quality targets is certainly also a challenge for conventional software~\cite{svensson2013investigation}, but training an ML model to go beyond a certain utility breakpoint turns into a functional requirement in practice. 
Andrew Ng, a prominent AI/ML advocate, recommends organizations to support this by defining a single-number evaluation metric for ML projects~\cite{ng2017machine}, thus making progress easier to measure.

\subsection{Explainability}
ML systems are hard to understand for human analysts since the decision algorithms are the result of a generic algorithm that is adjusted to fit specific data. Conventional programs are the results of a human developer who tries to transform her decision making strategy into code. Thus, program comprehension is crucial for developers to change the decision making procedures (i.e., the code). In ML systems, a developer usually does not (need to) change the code of the ML algorithm but instead manipulates the training data. This makes comprehension of ML systems challenging. 

Explainability was mentioned as important quality requirement in our interviews. P3: \enquote{Explainability is twofold: On the one side, there is a need to explain the model (what has been learned). On the other side, there is a need to explain single predictions of the model.}
The ability to explain decisions in ML systems depends on the used technology. While simpler ML techniques such as decision trees or Naive Bayes classifiers are easier to comprehend, it is harder, yet not impossible, to trace back decisions in more complex techniques such as Neural Networks~\cite{Ribeiro16,Winkler17}.

P4 mentioned that explainability may be even more important than predictive power: \enquote{Often, we constrain the models to derive explanations. We look for models that partition the input attributes to show relations between input and output. This decreases the predictive power but is usually favored by our customers.}
Another requirement related to explainability is simplicity of the models:
P1: \enquote{If there is a combination or transformation of features that is smaller but has a similar performance, we prefer that. [\ldots] We try to minimize the number of features to make the model more explainable.}

While the ML research community focuses on explaining trained models and specific decisions, there is no focus on specifying which situations demand an explanation. A requirements engineer should elicit explainability requirements from a user's point of view.

\subsection{Freedom from Discrimination}
ML systems are designed to discriminate. ML algorithms identify recurring patterns in data (i.e., stereotypes) and apply these patterns to judge about unseen data. However, some forms of discrimination are considered as unacceptable by our society or by law  (e.g., race or gender). Such features are illegal to use when working on systems that address insurance policies or filtering of job candidates. On the other hand, as explained by P1, gender and race might be essential in an ML-based decision support system for medical applications.

Freedom from discrimination is a new quality requirement not part of the established standards such as ISO\slash IEC 25010~\cite{ISO25010}. We describe \emph{freedom from discrimination} as \enquote{Using only logics of discrimination that are societally and legally accepted}. 
While discrimination is also possible in conventional rule-based algorithms, it is more critical in ML systems for two reasons: (1) Discrimination is more implicit in ML systems because you cannot pinpoint to the discriminating rule (e.g., \emph{if gender = f then health insurance rate += 20\%}), (2) ML algorithms amplify discrimination bias in the data during the training process (e.g., if a word is present in 60\% of training sentences, it might be predicted in 70\% of sentences at test time~\cite{Hendricks18}). P3 argued that a large fraction of discrimination in ML is introduced by unrepresentative training data, and indeed minority groups are in many settings underrepresented~\cite{barocas2018fairness}.

A requirements engineer must elicit and identify the ``protected'' characteristics that must not be used by the ML algorithm to discriminate samples. P2: ``There are tools and algorithms that can balance discrimination, but you have to know upfront which attributes you want to or must balance.''
Data scientists have two possibilities to ensure freedom from discrimination: (1) they can prepare the training data so that it does not contain any ``protected'' characteristics and (2) they can analyze the trained model to find important features in the training data. Afterward, a requirements engineer can assess whether these features may point to unacceptable discrimination. The possibility to analyze the trained model relates to the quality of explainability.  P1: \enquote{The question is where does discrimination occur? The answer lies in the explainability; features that play an important role. You can sort features by information gain. Then you can see if it is the religion that decides if you get a job or not.}


\subsection{Legal and Regulatory Requirements}

All interviewees brought up challenges regarding ethics and legal aspects. An example is how the General Data Protection Regulation (GDPR) constrains that personal data can only be used in ways specified by an explicit consent. P3 elaborated: ``The law states that you must know what your ML model will need to deliver before developing it. As a data scientist, I know that I probably need 20 out of 100 features but I don't know which 20. I have to ask for consent to use all of them. In the end, the legal department tells me that I'm collecting 80 features I'm not using. And that is illegal.'' Similarly, P2 highlighted that removing individual people from training data after revoked consent is non-trivial: ``When you're in the model, then it's not easy to remove you without retraining the entire model.'' P4, who is working in a highly regulated domain (finance), mentioned that their development process for ML systems must follow a predefined process to be accepted by regulatory authorities.  

We find it inevitable that requirements engineers working on ML systems must stay on top of legal requirements, and the data lineage must show that no illegal features have influenced the final dataset used for training the ML models.

\subsection{Data Requirements}
Training data is an integral part of any ML system. We envision that requirements for (training) data play a larger role for specifying ML systems than for conventional systems. We may even have \emph{data requirements} as a new class of requirements. P2: \enquote{It is a misconception when people talk about the \emph{algorithm} that does this or that. [\ldots] The difference [in ML systems] is that you always have to consider the data as well. [\ldots] Thus, the difference is that it is not only about the algorithm but also about the data.}
Breck et al.~\cite{Breck17} describe an analogy to code compilation: \enquote{One way to see this is to consider ML training as analogous to compilation, where the source is both code and training data. By that analogy, training data needs testing like code, and a trained ML model needs production practices like a binary does, such as debuggability, rollbacks and monitoring.} Based on our interviews, we would add \enquote{training data needs specified and validated requirements like code}. The ISO\slash IEC standard 25012~\cite{ISO25012} describes characteristics of data quality.
Interestingly, this standard is not as strongly used in RE as its sibling ISO\slash IEC 25010~\cite{ISO25010} on system quality characteristics. 
P3 mentioned that data scientists cannot specify requirements on the data itself: \enquote{You could try, but it won't help} -- the data is what it is, and it is up to the data scientist to make the most of it. 

\textbf{Data quantity:}
A general argument is \enquote{The model's performance was not good but we expect better performance if we have more data.} However, just \emph{more} data may not help. P2: \enquote{The more examples of a certain class you have in your data, the more characteristics you can learn. If you analyze a disease with many cases in your data, you find out a lot about the disease. If there is only one case with the disease, ML won't help.} That means, requirements on data quantity should not be specified on the number of examples but rather on the diversity of the examples.
P2 states: \enquote{The more diverse data you have, the more likely it is that outliers become classes on their own, which provide valuable information.} In some regulated domains, there are constraints on the amount of data necessary to tackle some problems. P4: \enquote{If we work on models to predict the likelihood of loan losses, we are forced to consider data from at least 5 years.}

P1 used to deal with small or homogeneous data: \enquote{In one case, we augmented the [small set of] training data with data from other sources. By this, we were able to explain phenomena in the original data.} 
We conclude that a requirements engineer should aim at identifying additional data sources as part of the stakeholder analysis. Stakeholders may have hypotheses about phenomena in the data. These hypotheses may point to additional data sources that could be used to augment the original data to get a richer set of training data.

\textbf{Data quality:}
There are several aspects of data quality that need to be addressed according to our interviewees. P1: \enquote{You have to check the data for \emph{garbage-in\slash garbage-out}. The higher the quality of the data, the better the application will work. That's why the process before the training is important: how I clean and augment the data.} P1 further refines the notion of data quality: \enquote{There are many dimensions of data quality. [\ldots] For me, the most important ones are completeness, consistency, and correctness}. Completeness refers to the sparsity of data within each characteristic (i.e., does the data cover the whole range of possible values). Consistency refers to the format and representation of data that should be the same in the dataset. Correctness refers to the degree to which you can rely on the data actually being true. Correctness is strongly influenced by the way how the data was collected. 

A common approach to meet the insatiable need for data in ML is to use public datasets, but P1 claims that they are often less trustable as ``nobody gets paid to maintain such data''. 
P2 stresses the risks related to data collected and labeled by humans: \enquote{If humans label data, you can almost be sure that your data is biased.} 
P3 has a general remark on how to lower the risk of incorrect data: \enquote{A common mistake is to use the collected data not only to train an ML model but also for other purposes.} 
P3 gave examples where training data was incorrect because the data was also used to monitor and control the people that reported the data.
The ML model appeared to perform well during development, but the results were useless when deployed. It turned out that the training data did not reflect reality, i.e., the data had been entered just to please the incentive system in the client organization. 

We conclude that an important activity of a requirements engineer is to identify and specify requirements regarding the collection of data, the data formats, and the ranges of data. This information needs to be elicited from the problem domain and serves as an input for data scientists. Requirements engineers must understand the importance of data provenance~\cite{wang2015big}, i.e., to critically question the data sources.

\section{RE Process for ML Systems} \label{sec:process}
In this section, we arrange our findings from the interviews according to the RE activities \emph{elicitation}, \emph{analysis}, \emph{specification}, and \emph{validation \& verification}. Note that we describe characteristics of these activities that are specifically important for ML systems. RE activities that are fundamental to any software engineering project are not addressed (e.g., understanding business goals and stakeholder needs). Moreover, we do not cover any ML feasibility analysis -- ML might not at all be an appropriate approach for the targeted solution. Table~\ref{tab:process-overview} summarizes our findings.

\begin{table*}
    \centering
    \caption{RE for ML: Impact on RE activities}
    \label{tab:process-overview}
    \begin{tabular}{@{}llll@{}}
    \toprule
    \textbf{Elicitation} & \textbf{Analysis}  & \textbf{Specification} & \textbf{V\&V}  \\
    \midrule
        \begin{minipage}[t]{0.2\textwidth}
            \begin{itemize}
            \setlength\itemsep{0.05cm}
            \item Elicit additional data sources
            \item Important stakeholders: Data scientists and legal experts
            \item ``Protected'' characteristics?
            \end{itemize}
        \end{minipage}&  
        \begin{minipage}[t]{0.25\textwidth}
            \begin{itemize}
            \setlength\itemsep{0.05cm}
            \item Discuss performance measures
            \item Discuss conditions for data preparation, definitions of outliers, and derived data 
            \end{itemize}
        \end{minipage}& 
        \begin{minipage}[t]{0.25\textwidth}
            \begin{itemize}
            \setlength\itemsep{0.05cm}
            \item Quantitative targets
            \item Data requirements
            \item Explainability
            \item Freedom from discrimination
            \item Legal and regulatory constraints
            \end{itemize}
        \end{minipage}& 
        \begin{minipage}[t]{0.22\textwidth}
            \begin{itemize}
            \setlength\itemsep{0.05cm}
            \item Analyze operational data
            \item Look for bias in data
            \item Retrain ML models
            \item Detect data anomalies
            \end{itemize}
        \end{minipage} \\
    \bottomrule
    \end{tabular}
\end{table*}

\subsection{Elicitation}
As mentioned in the interviews, ML systems usually profit from additional data that increases quantity and quality of the core data. As part of the stakeholder analysis, the requirements engineer should also identify all possibly relevant sources of data that may help the ML system to provide good and robust results.
Two important stakeholders for the development of ML systems are data scientists and legal experts. A data scientist should be consulted from the beginning to help eliciting requirements specific to data and its processing. Legal experts are required to determine constraints with respect to how the data is allowed to be used.
In the elicitation process, the requirements engineer should identify ``protected'' characteristics that must not be used by the ML algorithm to discriminate test samples. Similarly, the requirements engineer should elicit explainability requirements from a user's point of view (i.e., what are the situations or decisions that need to be explained to the user).

\subsection{Analysis}
Most important in requirements analysis for ML systems is to define and discuss the performance measures and expectations by which the ML system shall be assessed. As mentioned by our interviewees, performance measures such as accuracy, precision, or recall are not well understood by customers. On the other hand, the appropriateness of the measures depends on the problem domain. Therefore, a requirements engineer should bridge this gap by analyzing the demands of the stakeholders and translating them to appropriate measures. Thus, requirements engineers must have a good understanding of typical performance measures used for ML systems. A requirements engineer must explain the measures in the context of the domain (e.g., by explaining that a certain precision is an improvement given the a-priori distribution within the data). It is also important to analyze whether false positives and false negatives are equally bad. If this is not the case, the relevance of precision and recall may be different (cf.~\cite{Berry17,Winkler19}).  

During the so-called \emph{exploratory data analysis}, where the focus is on data understanding and data selection, a requirements engineer should facilitate the discussion between the customer and the data scientist. In that phase, data scientists decide which data to exclude, how to process and represent data, and how to augment data. 
P1 mentions that an early step in ML development is to understand the quality of the data (e.g., completeness, sparseness, and consistency) and how it must be enriched to constitute feasible training data. P2 discussed the tedious work of refining available data through various preprocessing steps. Preprocessing is an acquired skill in data science, the activity is only partly automated, and one must carefully maintain reproduceability or the resulting ML-systems will turn brittle. P2: \enquote{The data scientist tinkers a complex series of [preprocessing] steps together and there is an ML model. But when something stops working\ldots then it's bad.} The importance of the preprocessing means that requirements engineers must be able to understand prescriptive data lineage~\cite{miloslavskaya2016big}, i.e., the specification of explicit steps to process a dataset into its final state.
There is a risk that a data scientist makes decisions just because they improve the model but not based on the necessary domain understanding. A requirements engineer should support this task by eliciting, analyzing, and discussing conditions for data preparation, definitions of outliers, and derived data.

\subsection{Specification}
Requirements specifications for ML systems must have a stronger part on data requirements. A requirements engineer has to identify and specify requirements regarding the collection of data, the data formats, and the ranges of data. This information is elicited from the problem domain and serves as input for the data scientists, who analyzes the given dataset based on the requirements. Specifying data requirements includes information about the necessary quantity and quality of data. 

The requirements specification should also contain statements about the expected predictive power expressed in terms of the performance measures discussed during requirements elicitation and analysis. Performance on the training data can be specified as expected performance that can immediately be checked after the training process, whereas the performance at runtime (i.e., during operations) can only be expressed as desired performance that can only be assessed during operations.

Another focus in the specification should also be on the quality requirements pertinent to ML systems. The requirements engineer should specify whether discrimination is critical for the application (e.g., if it is people who are classified) and which characteristics should be ``protected'' (i.e., must not be used for classification). Similarly, the requirements engineer should specify explainability requirements that describe which situations and decisions of the ML system need to be explained to the user. Finally, the specification must contain regulations and constraints regarding the use of the data.

\subsection{Verification \& Validation}
Due to the dependency between the behavior of an ML system and the data it has been trained on, it is crucial to define actions that ensure that training data actually corresponds to real data. Since data characteristics in reality may change over time, requirements validation becomes an activity that needs to be performed continuously during system operation. Our interviewees agreed that monitoring and analysis of runtime data is essential for maintaining the performance of the ML system. They also agreed that ML systems need to be retrained regularly to adjust to recent data. By analyzing the problem domain, a requirements engineer should specify when and how often retraining is necessary. A requirements engineer should also specify conditions for data anomalies that may potentially lead to unreasonable behavior of the ML system during runtime. A checklist of measures to be considered during operations of ML systems is provided by Breck~et~al.~\cite{Breck17}. 
Apart from runtime monitoring, requirements validation also includes analyzing the training and production data for bias and imbalances. 

\section{Conclusions and Future Work}
In this paper, we made a first step towards exploring and defining an RE methodology for ML systems. We are convinced that RE for ML systems is special due to the different paradigm used to develop data-driven solutions. This methodology is not just a tailoring of a classical RE approach but includes new types of requirements such as explainability, freedom from discrimination, or specific legal requirements. The results of our study show that data scientists currently make many decisions with the goal to improve their ML models. For this task, they use and refer to technical concepts and measures that are often not well understood by the customer. There is a demand for requirements engineers who are able to base these decisions and technical concepts on a thorough understanding and analysis of the context. 

Data scientists have a large tool box of techniques at hand to balance, clean, validate, and explain data. However, it is the job of a requirements engineer to relate the application and results of such methods to the customers' needs and context. Existing data science guidelines such as~\cite{Breck17,Minhas19} can only address generic pitfalls and advices.

To broaden our results, we plan to extend our interview study and augment it with the view of requirements engineers of ML systems: Do they agree that RE for ML is different? If so, what do they do differently? If not, what are the reasons and implications? Additionally, we would like to augment our study with a third view from engineers of cyber-physical systems that are often safety-critical: Are there any ML-specific requirements that need to be included when ML is deployed in a safety-critical context?
Finally, we envision that systems in practice will never be pure ML systems. Therefore, it is an interesting question how to integrate RE for ML with the RE process for the surrounding (conventional) software system.  

\section*{Acknowledgment}
We thank the interview participants for their commitment and input. We thank our students for help with transcribing the interviews. 
This work was partly supported by the SMILE~II project financed by Vinnova, FFI, Fordonsstrategisk forskning och innovation under the grant number: 2017-03066.

\bibliographystyle{IEEEtran}
\bibliography{references}

\end{document}